\documentclass[letterpaper, 10 pt,conference]{ieeeconf} \IEEEoverridecommandlockouts
\usepackage{amsmath}
\usepackage{graphicx}
\usepackage[T1]{fontenc}
\usepackage{siunitx}
\usepackage{url}
\usepackage{lmodern}
\usepackage[english]{babel}
\usepackage{amsmath} 
\usepackage{mathtools}

\usepackage{amsfonts}
\usepackage{amssymb}
\usepackage{units}
\usepackage{afterpage}
\usepackage{gensymb}
\usepackage{color}
\usepackage{pgfplots}
\usepackage{algorithm}
\usepackage{algpseudocode}
\usepackage[capitalise]{cleveref}
\usepackage{microtype}
\pgfplotsset{compat=newest}
\pgfplotsset{plot coordinates/math parser=false,title style={yshift={-1.5mm}}}
\usepgfplotslibrary{groupplots}
\usetikzlibrary{plotmarks}
\usetikzlibrary{positioning}
\usetikzlibrary{shapes,arrows}
\usepgfplotslibrary{patchplots}
\definecolor{blue}{rgb}{0,0,0.55}
\definecolor{green}{rgb}{0,.9,0}
\definecolor{darkred}{rgb}{0.5,0,0}
\definecolor{yellow}{rgb}{.5,.5,0}
\definecolor{linecolor1}{rgb}{0,0,0.4}
\definecolor{linecolor2}{rgb}{1,0.55,0}
\definecolor{linecolor3}{rgb}{0.1,1,1}
\definecolor{linecolor4}{rgb}{0.6,0,0.5}

\newcommand{\figurecaptionreduction}{\vspace{-2.5mm}}

\newcommand{\T}{^{\hspace{-0.1mm}\scriptscriptstyle \mathsf{T}}\hspace{-0.2mm}}

\newcommand{\norm}[1]{\begin{Vmatrix}#1\end{Vmatrix}}
\newcommand{\inspace}[1]{\in \mathbb{R}^{#1}}

\newcommand{\w}{k}

\renewcommand{\vec}[1]{\operatorname{vec}{(#1)}}

\newcommand{\bmatrixx}[1]{\begin{bmatrix}#1\end{bmatrix}}

\newcommand{\set}[1]{\left\{ #1 \right\}}

\author{Fredrik Bagge Carlson\thanks{Open-source implementations of all presented methods and examples in this paper are made available at \cite{nn_prior}. The authors are members of the LCCC Linnaeus Center and the eLLIIT Excellence Center at Lund University, Dept Automatic Control, Lund Sweden\protect\\
{Fredrik.Bagge\_Carlson@control.lth.se}}  \qquad Rolf Johansson \qquad Anders Robertsson}

\title{\Large Tangent-Space Regularization for Neural-Network Models of Dynamical Systems}

\crefname{objective}{Objective}{Objectives}
\Crefname{objective}{Objective}{Objectives}
\crefname{algorithm}{Algorithm}{Algorithms}
\Crefname{algorithm}{Algorithm}{Algorithms}

\newlength\figureheight
\newlength\figurewidth
\begin{document}
\newtheorem{objective}{Learning Objective}

\maketitle

\begin{abstract}
  This work introduces the concept of tangent space regularization for neural-network models of dynamical systems. The tangent space to the dynamics function of many physical systems of interest in control applications exhibits useful properties, e.g., smoothness, motivating regularization of the model Jacobian along system trajectories using assumptions on the tangent space of the dynamics. Without assumptions, large amounts of training data are required for a neural network to learn the full non-linear dynamics without overfitting. We compare different network architectures on one-step prediction and simulation performance and investigate the propensity of different architectures to learn models with correct input-output Jacobian. Furthermore, the influence of $L_2$ weight regularization on the learned Jacobian eigenvalue spectrum, and hence system stability, is investigated.
\end{abstract}

\section{Introduction}
Dynamical control systems are often described in continuous time by differential state equations on the form
$$\dot x(t) = f_c\big(x(t), u(t)\big)$$
where $x$ is a Markovian state-vector, $u$ is the input and $f_c$ is a function that maps the current state and input to the state time-derivative. An example of such a model is a rigid-body dynamical model of a robot
\begin{equation}\label{eq:robot}
    \ddot x = -M^{-1}(x)  \big( C(x,\dot x)\dot x + G(x) + F(\dot x) - u \big)
\end{equation}
where $M,C,G$ and $F$ model phenomena such as inertia, Coriolis, gravity and friction.

In the discrete time domain, we may consider models on the form
\begin{equation}\label{eq:f}
    x_{t+1} = f\big(x_t, u_t\big)
\end{equation}
where $f$ is a function that maps the current state and input to the state at the next time-instance.
Learning a globally valid dynamics model $\hat{f}$ of an arbitrary non-linear system $f$ with little or no prior information is a challenging problem. Although in principle, any sufficiently complex function approximator, such as a deep neural network (DNN), could be employed, high demands are put on the amount of data required to prevent overfitting and obtain a faithful representation of the dynamics over the entire state space.

In many applications, the linearization of $f$ is important, a typical example being linear control design. Identification of $f$ must thus not only yield a good model for prediction/simulation, but also the Jacobian $J_{\hat f}$ of $\hat f$ must be close to the true system Jacobian $J$. In applications such as iterative learning control (ILC) and trajectory centric, episode-based reinforcement learning (TCRL), the linearization of the nonlinear dynamics along a trajectory is often all that is needed for optimization. Linear time-varying (LTV) models on the form
\begin{equation}
\label{eq:tvk1}
   x_{t+1} = A_t x_t + B_t u_t 
\end{equation}
where the matrices $A$ and $B$ constitute the output Jacobian,
 can be learned (data-)efficiently, in special cases extremely (time-)efficiently using dynamic programming \cite{ltvmodels}. A problem with learning an LTV model around a finite length trajectory is insufficient excitation provided by the control input. Prior knowledge regarding the evolution of the dynamics, encoded in form of carefully designed regularization, is required in order to obtain a well-posed optimization problem and a meaningful result. When model identification is a sub task in an outer algorithm that optimizes a control-signal trajectory or a feedback policy, excessive noise added for identification purposes may be undesirable, making regularization solely over time as in \cite{ltvmodels} insufficient.

 The discussion so far indicates two issues open for future work. 1) Complex non-linear models have the potential to be valid globally, but may suffer from overfitting and thus not learn a function with the correct linearization. 2) LTV models can be learned efficiently and can represent the linearized dynamics well, but are time-based and valid only locally.

 In this work, we make use of the regularization methods detailed in \cite{ltvmodels} for incremental learning of a general, non-linear black-box model, $\hat{f}$. We further develop a method to perform sampled tangent-space regularization for use with deep-learning frameworks without support for higher-order derivatives. To this end, we consider an episode-based setting where the dynamical model is updated after each episode. Each episode, a new trajectory $\tau_i = \set{x_t,u_t}_{t=1}^T$ is obtained, to which we fit an LTV model, $l_i$, on the form \labelcref{eq:tvk1} using the methods presented in~\cite{ltvmodels}\footnote{\url{github.com/baggepinnen/LTVModels.jl}} -- this model will provide the regularization for the non-linear model. We then update the non-linear state-space model $\hat f$ by adding $\tau_i$ to the set of training data for $\hat f$, while using $l_i$, which we assume have learned a good approximation of $J$ along $\tau_i$, for tangent-space regularization of $J_{\hat f}$.

 We proceed to introduce the problem of learning a dynamics model $\hat f$ in \cref{seq:f}. We then introduce tangent space regularization in \cref{seq:jacprop} and finally discuss the influence of weight decay on different formulations of the learning problem before conducting numerical evaluations.

 \section{Estimating the global model} \label{seq:f}

To frame the learning problem we let the dynamics of a system be described by a neural-network $\hat{f}$ to be fitted to input-output data $\tau$ according to
\begin{objective}\label{ob:1}
    \begin{align*}
        x_{t+1} &= \hat{f}(x_t,u_t) \inspace{n}\\
    \end{align*}
\end{objective}
which we will frequently write on the form $x^+ = \hat{f}(x,u)$ by omitting the time index $t$ and letting $\cdot^+$ indicate $\cdot_{t+1}$.
We further consider the linearization of $\hat{f}$ around a trajectory
\begin{equation}
\label{eq:tvk}
\begin{split}
   x_{t+1} &= A_t x_t + B_t u_t\\
   \w_t &= \vec{\bmatrixx{A_t\T & B_t\T}}
\end{split}
\end{equation}
where the matrices $A$ and $B$ constitute the input-output Jacobian $J_f$ of $f$
\begin{equation}
    J_f = \bmatrixx{\nabla_x f_1\T & \nabla_u f_1\T \\ \vdots & \vdots \\ \nabla_x f_n\T & \nabla_u f_n\T} \inspace{n \times (n+m)} = \bmatrixx{A & B}
\end{equation}
Our estimate $\hat f(x,u,w)$ of $f(x,u)$ will be parameterized by a vector $w$. The distinction between $f$ and $\hat f$ will, however, be omitted unless required for clarity.

We frame the learning problem as an optimization problem with the goal of adjusting the parameters $w$ of $\hat f$ to minimize a cost function $V(w)$ by means of gradient descent. The cost function $V(w)$ can take many forms, but we will limit the scope of this article to quadratic loss functions of the one-step prediction error, i.e.,
\begin{equation}
    V(w) = \dfrac{1}{2}\sum_t \big(x^+-\hat f(x,u,w)\big)\T\big(x^+-\hat f(x,u,w)\big)
\end{equation}

It is well known that the Jacobian of the discrete-time model $f$ has eigenvalues different from that of the continuous-time counterpart $f_c$. In particular, if the sample rate is high, most eigenvalues\footnote{We take the eigenvalues of a function to refer to the eigenvalues of the function Jacobian.} of $f_c$ are close to 0. The eigenvalues for the discrete-time $f$ however tend to cluster around 1 when sample rate is high. Another formulation of the discrete-time model, which we introduce as a new learning objective, is
\begin{objective}\label{ob:2}
    \begin{align*}
        x^+ - x &= \Delta x = g(x,u)\\
        f(x,u) &= g(x,u) + x
    \end{align*}
\end{objective}
with the second form being equivalent to the first, but highlighting a convenient implementation form that does not require transformation of the data. Classical theory for sampling of linear systems indicates that the eigenvalues of $g$ tend to those of $f_c$ as sample rate increases \cite{middleton1986delta}.

To gain insight into how this seemingly trivial change in representation may affect performance, we note that this transformation will alter the Jacobian according to
\begin{equation}
    J_g = \bmatrixx{A-I_n & B} \label{eq:jacshift}
\end{equation}
with a corresponding unit reduction of the eigenvalues of $A$. For systems with integrators, or slow dynamics in general, this transformation leads to a better conditioned estimation problem \cite{aastrom2013computer}. In~\cref{sec:results} we investigate whether or not this transformation leads to a better prediction/simulation result and whether modern neural network training techniques such as the ADAM optimizer \cite{kingma2014adam} and batch normalization \cite{ioffe2015batch} render this transformation superfluous. We further investigate how weight decay affects the eigenvalue spectrum of the Jacobian of $f$ and $g$ and hence, system stability.


\subsection{Optimization landscape}
To gain insight into the training of $f$ and $g$, we analyse the expressions for the gradients and Hessians of the cost functions.
For a linear model $x^+ = Ax+Bu$ and a least-squares cost-function $V(\theta) = \frac{1}{2}(y-\Phi \theta)\T(y-\Phi \theta)$, where the linear model is written on regressor form $y=\Phi \theta$ with all parameters of $A$ and $B$ concatenated into the vector $\theta$, the gradient and the Hessian are given by
\begin{align}
    \nabla_\theta V &= -\Phi\T(y-\Phi \theta)\\
    \nabla^2_\theta V &= \Phi\T\Phi
\end{align}
The Hessian is clearly independent of both the output $y$ and the parameters $\theta$ and differentiating the output does not have any major impact on gradient-based learning. For a nonlinear model, this is not necessarily the case:
\begin{align}
    V(w) &= \dfrac{1}{2}\sum_t \big(x^+-f(x,u,w)\big)\T\big(x^+-f(x,u,w)\big)\\
    \nabla_wV &= \sum_{t=1}^T \sum_{i=1}^n -\big(x_i^+-f_i(x,u,w)\big)\nabla_wf_i\\
    \nabla^2_wV &= \sum_{t=1}^T \sum_{i=1}^n \nabla_wf_i\nabla_wf_i\T -\big(x_i^+ -f_i(x,u,w)\big)\nabla^2_wf_i
\end{align}
In this case, the Hessian depends on both the parameters and the target $x^+$. The transformation from $f$ to $g$ changes the gradients and Hessians according to
\begin{align}
    \nabla_wV &= \sum_{t=1}^T \sum_{i=1}^n -\big(\Delta x_i-g_i(x,u,w)\big)\nabla_w(g_i)\\
    \nabla^2_wV &= \sum_{t=1}^T \sum_{i=1}^n \nabla_w g_i\nabla_w g_i\T -\big(\Delta x_i -g_i(x,u,w)\big)\nabla^2_wg_i
\end{align}
Since the output of $g$ is closer to zero compared to $f$ for systems of lowpass character, i.e., where $\norm{\Delta x}$ is small, the transformation can be seen as \emph{preconditioning} the problem by decreasing the influence of the term $\nabla^2_wg = \nabla^2_wf$ in the Hessian.\footnote{In the beginning of learning, the output of $\hat g$ is small due to the initialization of $w$.} With only the positive semi-definite term $\nabla_w(g)\nabla_w(g)\T = \nabla_w(f)\nabla_w(f)\T$ corresponding to $\Phi\T\Phi$ in the linear case remaining, the optimization problem becomes easier. Similarly, $g$ starts out closer to a critical point $\nabla_wV = 0$, making convergence rapid.

\section{Tangent-space regularization} \label{seq:jacprop}
For systems where the function $f$ is known to be smooth, the Jacobian $J_f(t)$ will vary slowly. In the rigid-body dynamical model~\labelcref{eq:robot}, for instance, the intertial and gravitational forces are changing smoothly with the joint configuration. A natural addition to the cost function of the optimization problem would thus be a tangent-space regularization term on the form
\begin{equation}\label{eq:reg}
    \sum_t\norm{\hat J_{t+1} - \hat J_t}
\end{equation}
which penalizes change in the input-output Jacobian of the model over time, a strategy we refer to as Jacobian propagation. A somewhat related strategy was proposed in~\cite{simard1998transformation}, where a tangent-dependent term in the cost function was successfully used to enforce invariance of an image classifier to rotations and translations of the input image.

Taking the gradient of terms depending on the model Jacobian requires calculation of higher order derivatives. Depending on the framework used for optimization, this can limit the applicability of the method. We thus proceed to describe a sampled version of tangent space regularization requiring only first order gradients.

 Estimation of an LTV model with the regularization term \labelcref{eq:reg} is straightforward and is the subject of~\cite{ltvmodels}. Given a good estimate $\hat J_f(t)$ provided by an LTV model, one may regularize the learning of $f$ by penalizing $\sum_t\norm{J_{\hat f}(t)- \hat J_f(t)}$. An approximation to Jacobian propagation is obtained by augmenting the dataset with input data perturbed in the direction of the Jacobian. For the regression task at hand, we can implement such a strategy by perturbing the input data $\left\{\tilde x,\tilde u\right\} = \left\{x+\epsilon_x,u+\epsilon_u\right\}$ by some small noise terms $\epsilon$, and generating a new target\footnote{New targets can be sampled each epoch.} $x^+$ using the LTV model $\tilde x^+ = A\tilde x + B \tilde u$. If this was done for each component of $x$ and $u$ separately, this would correspond exactly to finite-difference Jacobian propagation. However, due to the smoothness assumption of $f$ together with the smooth inductive bias of neural networks, we get reasonable results by only perturbing each input instance by a small Gaussian random vector. We refer to this strategy as \emph{sampled} Jacobian propagation. \Cref{sec:results} demonstrates how this approach enables learning of neural-network models with high-fidelity Jacobians. The procedure is summarized in~\cref{alg:composite}.
\begin{algorithm}[t]
    \caption{An algorithm for efficient learning of a nonlinear neural-network dynamics model. The sampling of a rollout may entail using the estimated models for, e.g., iterative learning control, trajectory optimization or reinforcement learning.}
    \label{alg:composite}
    \small
    \begin{algorithmic}
        \State Initialize a model $\hat f$.
        \Loop
        \State Sample rollout trajectory $\tau_i = \set{x_t,u_t}_{t=1}^T$
        \State Fit LTV model $l_i$ using method in~\cite{ltvmodels}
        \State Generate perturbed trajectories $\tilde\tau_i = \set{\tilde x_t,\tilde u_t}$ using $l_i$.
        \State Fit $\hat f$ to $\tau_i$ and $\tilde\tau_i$
        \State (Optimize controller or control signal trajectory)
        \EndLoop
    \end{algorithmic}
\end{algorithm}

\section{Weight decay}
Weight decay is commonly an integral part of model fitting used to combat overfitting, and can be thought of as either penalizing complexity of the model or as encoding prior knowledge about the size of the model coefficients. $L_2$ weight decay is, however, a blunt weapon. While often effective at mitigating overfitting, the procedure might introduce a bias in the estimate. Since the bias always is directed towards smaller weights, it can have different consequences depending on what small weights imply for a particular model architecture. For a discrete-time model $f$, small weights imply small eigenvalues and a small output. For $x^+ = g(x,u) + x$, on the other hand, small weights imply eigenvalues closer to 1. Weight decay thus has a vastly different effect on learning $f$ and $g$, and since a high sample rate implies eigenvalues closer to 1, weight decay is likely to bias the result of learning $g$ in a milder way as compared to when learning $f$. We explore the effect of weight decay in the next section by fitting models to data generated by linear systems.

A natural question to ask is if weight decay can bias the eigenvalues of the learned function to arbitrary chosen locations. A generalized form of model formulation is
    $x^+ - Ax - Bu = h(x,u)$
where $A$ and $B$ can be seen as a nominal linear model around which we learn the nonlinear behavior. Weight decay will for this formulation bias the Jacobian towards $A$ and $B$ which can be chosen arbitrarily. Obviously, choosing a nominal linear model is not always easy, and may in some cases not make sense. One can however limit the scope to formulations like $x^+ - \tau x = h(x,u)$, where $\tau$ is a scalar or a diagonal matrix that shifts the nominal eigenvalues along the real axis to, e.g., encourage stability.

\section{Evaluation}\label{sec:results}
The previously described methods were evaluated on two benchmarks problems. We compared performance on one-step prediction as well as simulation from a known initial condition. We further compared the fidelity of the Jacobian of the estimated models, an important property for optimization algorithms making use of the models. The benchmarks consist of a simulated robot arm with two revolute joints, friction and gravity and randomized, stable linear systems. Code to reproduce results are presented in the online repository.

We initially describe a baseline neural network approximator used in the experimental evaluation which we use to draw conclusions regarding the different learning objectives. We describe how deviations from this baseline model alter the conclusions drawn in Appendix~\labelcref{sec:deviations}. The first examples demonstrate the effectiveness of tangent-space regularization, whereas later examples demonstrate the influence of weight decay.

\subsection{Nominal model}

Both functions $f$ and $g$ are modeled as ensembles, $\mathcal{E}$, of 4 distinct single-layer neural networks with 20 neurons each. The 4 networks in the ensemble are trained on the same data with the same objective function, but with different random initializations of the weights and different non-linear activation functions.\footnote{The variation in the ensemble predictions may serve as a bootstrap estimate of the variance of the estimator.} After a comparative study of 6 different activation functions, presented in Appendix~\labelcref{sec:activations}, we selected networks with the activation functions elu, sigmoid, tanh and swish to form the components of the ensemble $\mathcal{E}$. We train the models using the ADAM optimizer with a fixed step-size and fixed number of epochs, 2000 for $f$ and 1000 for $g$. The entire framework including all simulation experiments reported in this article is published at \cite{nn_prior} and is implemented in the Julia programming language~\cite{bezanson2017julia} and the Flux machine learning library.\footnote{\url{github.com/FluxML/Flux.jl}}

When training with tangent-space regularization, a single perturbed trajectory was added with $$\left\{\epsilon_x,\epsilon_u\right\} \sim \mathcal{N}\big(0, \left\{0.1^2\operatorname{diag}{\Sigma_x},\; 0.1^2\operatorname{diag}{\Sigma_u}\right\}\big)$$
and the network was trained for half the number of epochs compared to the nominal baseline model.

\subsection{Randomized linear system}
\begin{figure}[htp]
    \centering
    \includegraphics[width=\linewidth]{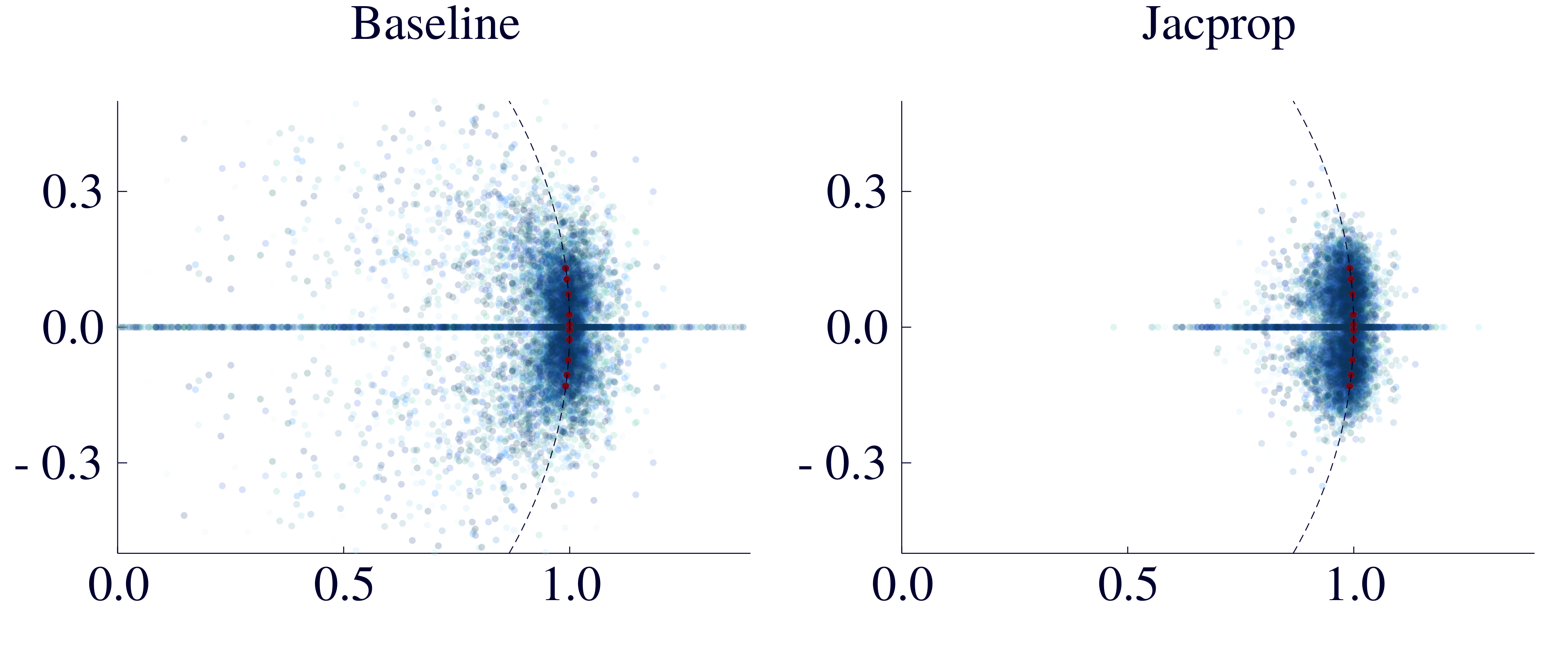}
    \caption{Learned Jacobian eigenvalues of $g$ for randomly sampled points in the state-space (blue) together with the eigenvalues of the true model (red). Tangent-space regularization (right) leads to better estimation of the Jacobian with eigenvalues in a tighter cluster around the true eigenvalues close to the unit circle.}
    \label{fig:jacpropeig}
\end{figure}

To assess the effectiveness of sampled Jacobian propagation we create random, stable linear systems according to~\cref{alg:randomlinear} and evaluate the Jacobian of the learned model for points sampled randomly in the state space. The results are illustrated in~\cref{fig:jacpropeig}. During training, the model trained without tangent-space regularization reaches far lower training error, but validation data indicate that overfitting has occurred. The model trained with tangent-space regularization learns better Jacobians and produces smaller prediction- and simulation errors.

\begin{algorithm}[t]
    \caption{Generation of random, stable linear systems.}
    \label{alg:randomlinear}
    \small
    \begin{algorithmic}
        \State $A_0 = 10 \times 10$ matrix of random coefficients
        \State $A = A_0-A_0\T$        skew-symmetric = pure imaginary eigenvalues
        \State $A = A - \Delta t \, I $    make 'slightly' stable
        \State $A = \exp(\Delta t \, A)$   discrete time, sample time $\Delta t$
        \State $B =$ random coefficients
    \end{algorithmic}
\end{algorithm}

\begin{figure}[htp]
    \centering
    \includegraphics[width=\linewidth]{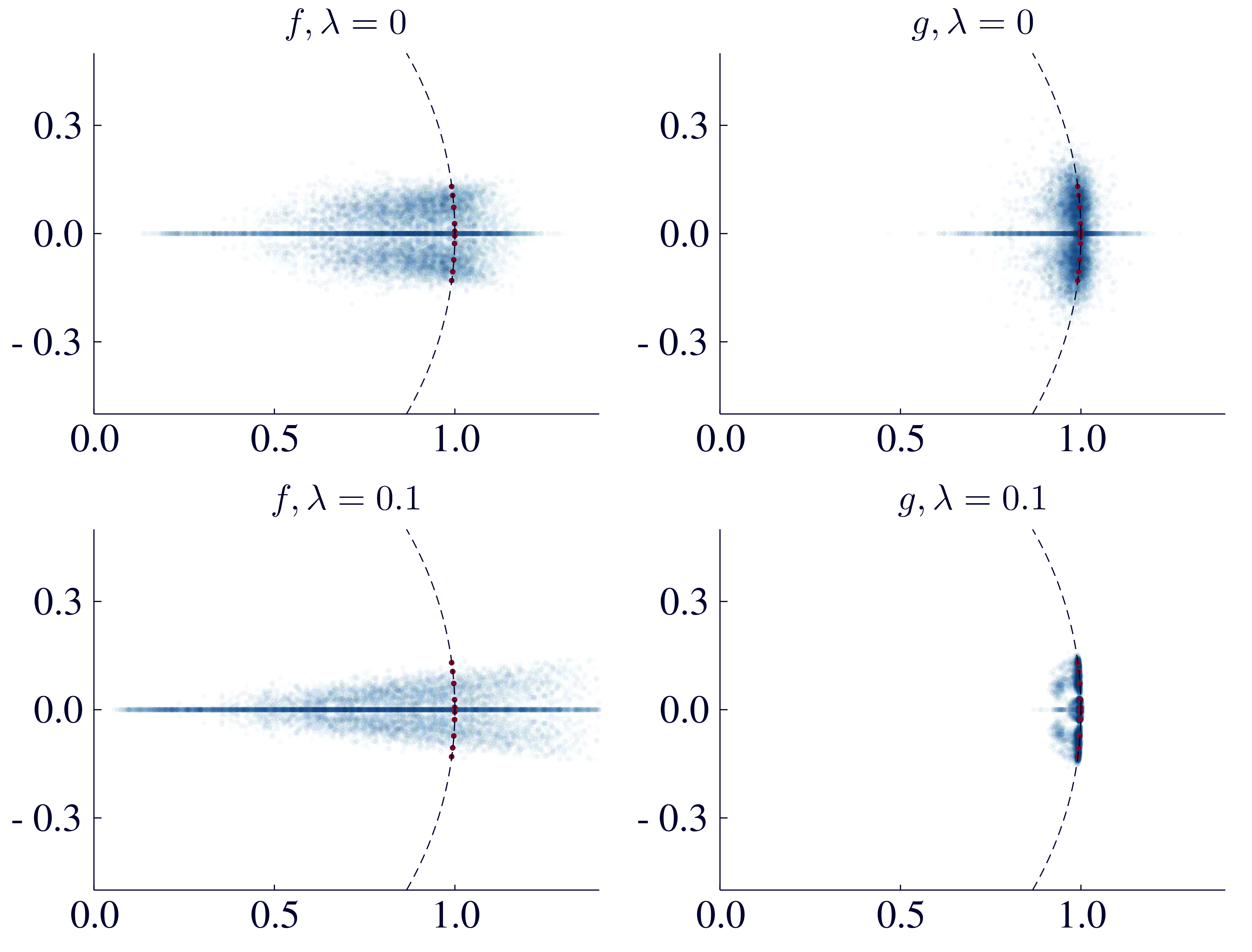}
    \caption{Eigenvalues of learned Jacobians for the linear system task. The top/bottom rows shows models trained without/with weight decay, left/right columns show $f$/$g$. Weight decay has deteriorating effect on learning $f$, pulling the eigenvalues towards 0, while being beneficial for learning $g$, keeping the eigenvalues close to 1 on the unit circle. All models are trained with sampled Jacobian propagration.}
    \label{fig:wd}
\end{figure}

The effect of weight decay on the learned Jacobian is illustrated in~\cref{fig:wd}. Due to overparamterization, heavy overfitting is expected without adequate regularization. Not only is it clear that learning of $g$ has been more successful than learning of $f$ in the absence of weight decay, but we also see that weight decay has had a deteriorating effect on learning $f$, whereas it has been beneficial in learning $g$. This indicates that the choice of architecture interacts with the use of standard regularization techniques and must be considered while modeling.

\subsection{Robot task}
The robot has non-linear dynamics and thus a changing Jacobian\footnote{We are referring to the Jacobian of the dynamics, as opposed to the Jacobian of the forward kinematic encountered in robotics.} along a trajectory. This task demonstrates the utility of tangent-space regularization for systems where the regularization term is not the theoretically perfect choice, as was the case with the linear system. We simulate the system with a low-pass filtered random input and compare prediction and simulation error as well as the error in the estimated Jacobian.

\begin{figure}
    \centering
    \setlength{\figurewidth}{0.45\linewidth}
    \setlength{\figureheight}{4cm}
    \input{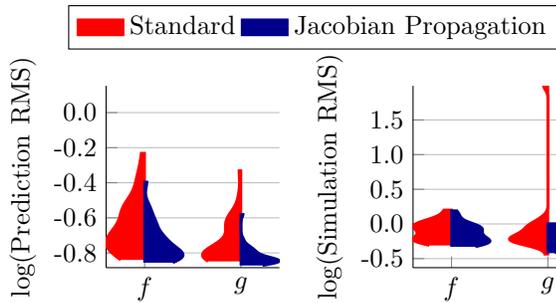}
    \figurecaptionreduction
    \caption{Distribution of prediction and simulation errors on the validation data for the robot task. Each violin represents 35 Monte-Carlo runs. The figure indicates that tangent-space regularization through tangent-space regularization stabilizes learning and reduces both one-step prediction error and simulation error.}
    \label{fig:valerr}
\end{figure}
The prediction and simulation errors for validation data, i.e., trajectories not seen during training, are shown in~\cref{fig:valerr}. The results indicate that tangent-space regularization leads to reduced prediction and simulation errors compared to baseline, with lower mean error and smaller spread, indicating more robust learning. In particular, divergence of the state was encountered during some simulations of the baseline model, a behavior which was not encountered for models trained with tangent-space regularization.

\begin{figure}
    \centering
    \setlength{\figurewidth}{0.99\linewidth}
    \setlength{\figureheight }{5.7cm }
    \input{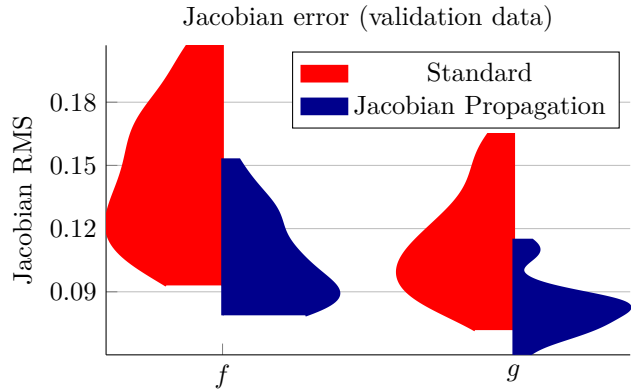}
    \figurecaptionreduction
    \caption{Distribution of errors in estimated Jacobians (Eq.~(\ref{eq:jacerr})). Each violin represents 35 Monte-Carlo runs. Networks trained with tangent-space regularization exhibit significantly less error in the estimated Jacobians compared to networks trained the conventional way.}
    \label{fig:jacerr}
\end{figure}

To assess the fidelity of the learned Jacobian, we average the input-output Jacobian over the ensemble $\mathcal{E}$ and compare it to the ground-truth Jacobian of the simulator. We display the distribution of errors in the estimated Jacobians over 35 Monte-Carlo runs\footnote{The number was chosen based on the number of cores available.} in~\cref{fig:jacerr}. The error was calculated as the mean over time steps of the Frobenius norm of the difference in coefficients between the true Jacobian and the ensemble average Jacobian:
\begin{equation}\label{eq:jacerr}
    \sqrt{\dfrac{1}{T}\sum_t \norm{J_t - \dfrac{1}{|\mathcal{E}|}\sum_{\mathcal{E}}\hat J_t}_F^2}
\end{equation}
The results show a significant benefit of tangent-space regularization over baseline, with a reduction of the mean error as well as a smaller spread of errors, indicating a more stable and reliable training.

\section{Discussion}

We note that $g$ generally trains faster and reaches a lower value of the objective function compared to $f$. The structure of $g$ resembles that of a residual network \cite{he2016deep}, where a \emph{skip connection} is added between the input and a layer beyond the first adjacent layer, in our case, directly to the output. While skip connections have helped to enable successful training of very deep architectures for tasks such as image recognition, we motivated the benefit of the skip connection with classical theory for sampling of continuous-time systems~\cite{middleton1986delta} and an analysis of the model Hessian. Exploring the similarities with residual networks remains an interesting avenue for future work.

The scope of this article was limited to settings where a state-sequence is known. In a more general setting, learning the transformation of past measurements and inputs to a state-representation is required and networks with recurrence (RNNs) are required. Initial results indicate that the conclusions drawn regarding the formulation ($f$ vs. $g$) of the model and the effect of weight decay remains valid in the RNN setting, but a more detailed analysis is the target of future work.

\begin{figure*}[htb]
    \centering
    \setlength{\figurewidth}{0.99\linewidth}
    \setlength{\figureheight }{3.5cm}
    \input{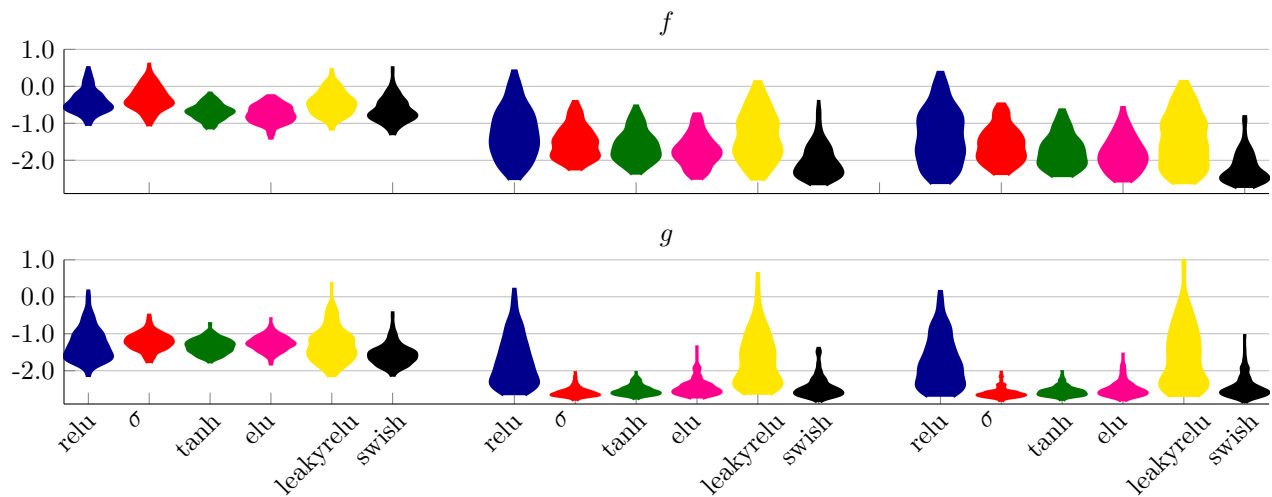}
    \figurecaptionreduction
    \caption{Distributions (cropped at extreme values) of log-prediction errors on the validation data after 20, 500 and 1500 (left, middle, right) epochs of training for different activation functions. Every violin is representing 200 Monte-Carlo runs and is independently normalized such that the width is proportional to the density, with a fixed maximum width.}
    \label{fig:activations}
\end{figure*}

\section{Conclusions}

We have demonstrated how tangent-space regularization by means of sampled Jacobian propagation can be used to regularize the learning of a neural network model of a dynamical system with an increase in prediction and simulation performance as well as increasing the fidelity of the learned Jacobians as result.

We investigated different architectures of the NN model and found that the relationship between sample time and system bandwidth affects the preferred choice of architecture, where one approximator architecture ($g$) train faster and generally generalize better in terms of all metrics if the sample rate is high. An analysis of gradient and Hessian expressions motivated the difference and conclusions were reinforced by experiment with different sample rates.

The effect of including $L_2$ weight decay was investigated and shown to vary greatly with the model architecture. Implications on the stability and eigenvalues of the learned model highlights the need to consider the architecture choice carefully.

\appendices
\section{Comparison of activation functions}\label{sec:activations}
\Cref{fig:activations} displays the distribution of prediction errors over 200 Monte-Carlo runs with different random seeds and different activation functions. The results indicate that the relu and leaky relu functions are worse suited for the considered task and are thus left out from the set of selected bootstrap ensemble activation functions.

\section{Deviations from the nominal model}\label{sec:deviations}

\subsubsection*{Sample rate} The sample rate determines the location of the eigenvalues of the Jacobian for a discrete-time dynamical system. Faster sampling moves the eigenvalues closer to 1, with implications for numerical accuracy for very small sampling times \cite{aastrom2013computer}. Faster sampling also leads to a function $g$ closer to zero and $f$ closer to 1.
With 5 times slower sampling, the difference between $f$ and $g$ in prediction performance was less pronounced for the validation data and both methods performed comparably with respect to the estimated Jacobians. With 5 times faster sampling, method $g$ performs much better on prediction and Jacobian estimation but much worse on simulation.
\subsubsection*{Number of neurons} Doubling or halving the number of neurons generally led to worse performance.
\subsubsection*{Number of layers} Adding a fully connected layer with the same number of neurons did not change any conclusions.
\subsubsection*{Dropout} Inclusion of dropout (20\%) increases prediction and simulation performance for $g$ while performance is decreased for $f$. Performance on Jacobian estimation was in general worse.
\subsubsection*{Layer normalization} Prediction and simulation performance remained similar or slightly worse, whereas Jacobian estimation broke down completely.
\subsubsection*{Measurement noise} Although not a hyper parameter, we investigate the influence of measurement noise on the identification results. Increasing degree of measurement noise degrades all performance metrics in predictable ways and does not change any qualitative conclusions.
\subsubsection*{Ensemble size} A three times larger ensemble of network yields marginally better performance on all metrics, with a three times longer training time.
\subsubsection*{Number of perturbed trajectories for sampled Jacobian propagation} We found that sampling a single trajectory was sufficient and adding additional samples did not alter the performance significantly. A larger state dimension might require more samples, although we have not quantified this.
\subsubsection*{Optimizer, Batch normalization, Batch size} Neither of these modifications altered the performance significantly.

\bibliography{bibtexfile}{}
\bibliographystyle{IEEEtran}
\end{document}